%% file: main.tex
\pdfoutput=1
\documentclass[11pt]{article}

\usepackage{EMNLP2023}

\usepackage{times}
\usepackage{latexsym}

\usepackage[T1]{fontenc}
\usepackage[utf8]{inputenc}

\usepackage{microtype}

\usepackage{inconsolata}

\usepackage{amssymb}
\usepackage{amsmath}
\usepackage{amsfonts}
\usepackage{graphicx}
\usepackage{threeparttable}
\usepackage{booktabs}
\usepackage{xspace}
\usepackage{url}

\usepackage{listings}
\definecolor{codegreen}{rgb}{0,0.6,0}
\definecolor{codegray}{rgb}{0.5,0.5,0.5}
\definecolor{codepurple}{rgb}{0.58,0,0.82}
\definecolor{backcolour}{rgb}{0.95,0.95,0.92}
\lstdefinestyle{mystyle}{
    backgroundcolor=\color{backcolour},   
    commentstyle=\color{codegreen},
    keywordstyle=\color{magenta},
    numberstyle=\tiny\color{codegray},
    stringstyle=\color{codepurple},
    basicstyle=\ttfamily\footnotesize,
    breakatwhitespace=false,         
    breaklines=true,                 
    captionpos=b,                    
    keepspaces=true,                 
    numbers=left,                    
    numbersep=5pt,                  
    showspaces=false,                
    showstringspaces=false,
    showtabs=false,                  
    tabsize=2
}
\newcommand{\effocr}{\texttt{EffOCR}\xspace}

\begin{document}

%%%%%%%%% TITLE - PLEASE UPDATE
\title{\effocr: An Extensible, Open-Source Package for Efficiently Digitizing World Knowledge}
\author{Tom Bryan$^{1}$, Jacob Carlson$^{1}$, Abhishek Arora$^{1}$, Melissa Dell$^{1, 2\ast}$ \\
\normalsize{$^{1}$ Harvard University; Cambridge, MA, USA.}\\
\normalsize{$^{2}$ National Bureau of Economic Research; Cambridge, MA, USA.}\\
\normalsize{All authors contributed equally.}
\normalsize{$^\ast$Corresponding author:  melissadell@fas.harvard.edu.}
}
\maketitle

%%%%%%%%% ABSTRACT
\begin{abstract}
     Billions of public domain documents remain trapped in hard copy or lack an accurate digitization. Modern natural language processing methods cannot be used to index, retrieve, and summarize their texts; conduct computational textual analyses; or extract information for statistical analyses, and these texts cannot be incorporated into language model training. Given the diversity and sheer quantity of public domain texts, liberating them at scale requires optical character recognition (OCR) that is accurate, extremely cheap to deploy, and sample-efficient to customize to novel collections, languages, and character sets. Existing OCR engines, largely designed for small-scale commercial applications in high resource languages, often fall short of these requirements. \effocr (\texttt{\textbf{Eff}icient\textbf{OCR}}), a novel open-source OCR package, meets both the computational and sample efficiency requirements for liberating texts at scale by abandoning the sequence-to-sequence architecture typically used for OCR, which takes representations from a learned vision model as inputs to a learned language model. Instead, \effocr models OCR as a character or word-level image retrieval problem. \effocr is cheap and sample efficient to train, as the model only needs to learn characters' visual appearance and not how they are used in sequence to form language. Models in the \effocr model zoo can be deployed off-the-shelf with only a few lines of code and include lightweight models designed for mobile phones that are extremely cheap to deploy. Importantly, \effocr also allows for easy, sample efficient customization with a simple model training interface and minimal labeling requirements due to its sample efficiency.  We illustrate the utility of \effocr by cheaply and accurately digitizing 20 million historical U.S. newspaper scans, evaluating zero-shot performance on randomly selected documents from the U.S. National Archives, and accurately digitizing a Japanese document collection for which all other OCR solutions failed. 
\end{abstract}

%%%%%%%%% BODY TEXT
\section{Introduction}
\label{sec:intro}

Vast document collections remain trapped in hard copy or lack accurately digitized texts. For example, the U.S. National Archives holds approximately 13.28 billion pages of textual records, most of which are in the public domain.\footnote{For documents published in the United States, the public domain includes any content published by a U.S. government officer/employee in the course of official duties, all content published more than 95 years ago, and some content published before 1989 that either wasn't published with a notice or did not renew copyright. This is common, for instance, in the case of publications like local newspapers \cite{copyright}. See the supplementary materials for details.} These documents are preserved because they are central to the workings of the U.S. government, have long-term research value, or provide valuable information for the public, but working with most of them is costly and time-consuming. The U.S. National Archives are not unique: many other countries have national archives with public domain collections numbering in the billions of pages, not to mention state and local archives and libraries. Without accurate machine-readable data, modern natural language processing (NLP) tools cannot be used to index, retrieve, and summarize materials; conduct computational textual analyses; or extract information for statistical investigations. Public domain texts, if accurately digitized, could also provide massive scale information for training large language models, with no risks of copyright infringement. 

Using optical character recognition (OCR) to digitize public domain collections on a large scale entails several challenges. 

\textbf{Cost:} First, the OCR solution must be cheap to deploy, given document collections whose size numbers in the millions or even billions of pages. Commercial engines - as well as large open-source OCR models - fall well short of this requirement. Using them to digitize large-scale collections would require astronomical budgets. 

\textbf{Accuracy:} Second, digitized texts need to be sufficiently accurate for end users' objectives, which are highly diverse. 
%Some popular OCR engines that can be deployed cheaply rely on older architectures, \textit{e.g.,} an LSTM language model \cite{EasyOCR, tesseract}, and tend to fall short of the accuracy of state-of-the-art models. 
Accuracy can be particularly central for quantitative applications, for which small errors can create major statistical outliers. Models for lower resource languages, if they exist, tend to perform much worse than models for high resource settings like English.

\textbf{Sample efficient, easy training:}
Documents are highly heterogeneous in terms of their fonts or handwritings, languages, scripts, backgrounds, and artifacts from scanning and aging. There are a diversity of documents for which no existing OCR solution works zero-shot, particularly in low resource languages. Yet stakeholders who would like to digitize these documents rarely have familiarity with deep learning frameworks. Bringing high quality OCR to low resource settings requires a simple API for training and a sample efficient architecture, with an accessible compute and annotation burden.

\textbf{A diversity of pre-trained and tuneable models:}
Users have diverse accuracy needs, scaling requirements, and budgets. A comprehensive OCR solution would make it easy to compare the accuracy and deployment costs of models of varying sizes so that users can choose the one that best suits their needs for a particular application. 

To meet these objectives, we developed \effocr, an open-source OCR package designed for researchers, libraries, and archives seeking a computationally and sample efficient OCR solution for digitizing diverse document collections. EffOCR has two key ingredients: 1) a novel OCR architecture and 2) a carefully designed interface to facilitate off-the-shelf OCR usage, customization via model training if necessary, and easy sharing of OCR models.  

The novel \effocr model architecture is treated in detail in \citet{carlson2023efficient}, where we compare accuracy, sample efficiency, and deployment costs to a range of popular OCR engines. 
In short, OCR predominantly models text recognition as a sequence-to-sequence (seq2seq) problem, in which learned representations from a vision model are taken as inputs to a learned language model. Learning how vision embeddings are used in sequence to form language requires large amounts of data. For example, the predominant transformer sequence-to-sequence OCR package \cite{li2021trocr} was trained on 684 million text lines using 32 32GB V100 GPU cards. 
State-of-the-art seq2seq OCR is sample-inefficient to tune and infeasible for users to extend to low resource languages, which may not even have a transformer large language model (LLM) that can be used to initialize the model, as language modeling advances are concentrated in less than two dozen languages  \cite{joshi2020state}.
The typical stakeholder working with low resource documents has a minimal budget for training and limited experience with deep learning frameworks, underscoring the need for a much more sample efficient framework with an easy-to-use API. 

Additionally, seq2seq OCR requires autoregressive decoding, which makes inference slower than it would be, all else equal, with parallel decoding. 

\effocr abandons the seq2seq OCR model that predominates in the literature, instead modeling OCR as a word or character level \textit{image} retrieval problem. \effocr first localizes words using highly accurate, scalable object detection methods \cite{yolov8, chen2019mmdetection, wu2019detectron2}. Recognition is then modeled as a contrastively trained image retrieval problem, where image embeddings of the same character or word have similar representations, regardless of their style. \effocr is trained primarily on digital fonts, combined with a modest number of character and word crops from real-world documents. At inference time, characters/words are recognized by computing their nearest neighbor in an offline dictionary of exemplar embeddings created with a digital font.  %Not all words will appear in a dictionary - whether due to hyphenation at the end of a line, proper nouns, or acronyms - and so when a word's representation is below a threshold similarity to the nearest word representation in the index, we default to localizing characters and employing character level image retrieval. Relative to the sequence-to-sequence architecture, 
%\effocr is akin to how humans read texts: recognizing words by sight and ``sounding out'' unrecognized words character-by-character.  
\citet{carlson2023efficient} show, using English, Japanese, and Polytonic Greek benchmarks, that the \effocr architecture is accurate, highly sample efficient, cheap to train, and extremely fast to deploy when using backbones designed for mobile phones. %They show strong zero shot performance on a random sampling of documents from the U.S. National Archives, on historical U.S. patent data, and on historical newspapers that were not seen during training. 

To meet the challenges of digitizing large-scale and low-resource document collections, the \effocr package contains the following components: 
\begin{enumerate}
    \item An off-the-shelf toolkit for applying OCR models with just a few lines of code
    \item A repository of pre-trained OCR models that underlies off-the-shelf usage
    \item ONNX runtime support for fast deployment
    \item Comprehensive tools for efficient model tuning 
     \item Supports models from popular backends \cite{chen2019mmdetection, yolov8} for initializing localization and any \textit{timm}-supported model for initializing recognition
    \item Easy sharing of models, to promote reusability, reproducibility, and extensibility 
\end{enumerate}

\effocr has been extensively tested. For example, we have used it to cheaply digitize 20 million pages of historical public domain U.S. newspaper scans that are extremely heterogeneous, posting the massive-scale output to Hugging Face.\footnote{\url{https://huggingface.co/datasets/dell-research-harvard/AmericanStories}}
Creating this dataset within our modest budget while meeting accuracy requirements would have been impossible without \effocr.
We have also examined performance in settings where no existing OCR solutions provide usable output, and tested zero-shot performance on a random selection of U.S. National Archive documents, with a model that did not see any similar content during training. 
%A demo of \effocr is available here: \url{https://www.youtube.com/watch?v=2rPMVbim9t4}. 
Tutorials are available at \url{https://effocr.github.io/}.

\effocr has a GNU General Public License. 
It is being actively maintained and crowd-sourcing of annotations to expand the pre-trained model zoo to other languages and settings, including handwriting, is underway.

The rest of this paper is organized as follows. Section \ref{lit} briefly compares \effocr to existing, popular OCR solutions. Section \ref{sec:core-library} describes the key features of the OCR package, and Section \ref{sec:use-case} examines several use cases: using \effocr to digitize 20 million historical newspaper scans, using \effocr zero-shot on randomly selected collections from the U.S. National Archives, and using \effocr to digitize a historical Japanese publication for which all existing OCR solutions fail. Finally, Section \ref{limits} discusses the limitations of the \effocr package. 

\section{Comparisons to Other OCR Engines} \label{lit}
There is a vast literature on OCR. Of primary interest here are widely used OCR softwares, which are the most plausible alternatives to \effocr. 

\effocr - as the name suggests - is tailored towards applications requiring computational or sample efficiency.
\citet{carlson2023efficient} conduct detailed experiments comparing the EffOCR architecture to other widely used solutions, considering accuracy, sample efficiency, and computational efficiency. We refer the interested reader to that paper for details, summarizing the two key themes that emerge here.

\textbf{Customization is highly relevant:} As the preponderance of researchers still using data entry firms suggests, sometimes no existing OCR solution provides acceptable accuracy.  For typewritten Japanese documents from the mid-20th century, that are of considerable relevance to studying Japan's remarkable 20th century growth performance, \citet{carlson2023efficient} show that the best performing engine (Baidu, the leading commercial OCR for Asian languages) gets over half of characters wrong. The widespread failure of OCR to provide acceptable results is also evidenced by a large post-OCR error correction literature (\textit{e.g.}, \citet{10.1162/tacl_a_00379, 10.1145/3453476, artidigh20}). 

\effocr is significantly more sample efficient than leading open-source OCR engines: EasyOCR \cite{EasyOCR}, TrOCR \cite{li2021trocr}, and PaddleOCR \cite{du2022svtr}, as shown in the supplementary materials.\footnote{EasyOCR uses a seq2seq convolutional recurrent neural network (CRNN) framework \cite{shi2016end}, TrOCR uses a seq2seq encoder-decoder transformer \cite{li2021trocr}, and PaddleOCR's uses Single Vision Text Recognition (SVTR), which like \effocr abandons seq2seq, dividing text images into small (non-character) patches, using mixing blocks to perceive inter- and intra-character patterns, and recognizing text by linear prediction \cite{du2022svtr}.}
Learning to recognize the visual features of individual characters is a highly parsimonious problem, making \effocr cheap to tune or train from scratch. 
%Moreover, in some contexts like digitizing tables (common in quantitative social sciences), there may be little language context in any case.
Because \effocr does not need to understand language, it is straightforward to extend to new languages and scripts, including those that lack a transformer large language model to initialize state-of-the-art seq2seq. 
The convolutional models in the \effocr model zoo can be trained on a Google Colab account, whereas training TrOCR on 684 million text lines required 32 32GB V100 cards. 

A central aim of \effocr is to democratize access to OCR to low resource languages and settings that are difficult to study because existing solutions are not suitable to these use cases. While we do not have the resources to train OCR models for all these settings, our simple APIs for training models and uploading them to the \effocr model hub can encourage the crowdsourcing of this effort. 

\textbf{The most accurate OCR engines in high resource settings (\textit{e.g.,} English) are costly to deploy at scale:} TrOCR (Base) is a highly accurate state-of-the-art English OCR. With 334 million parameters, it is nearly 50 times slower to deploy than our pre-trained lightweight \effocr English word recognition model, while offering only relatively modest gains on the evaluation tasks in \citet{carlson2023efficient}.\footnote{TrOCR has a small model (62M parameters), but \citet{carlson2023efficient} find it is outperformed by the 334M parameter base model by a wide margin on historical documents.}  For English, Google Cloud Vision (GCV) - a proprietary commercial product - dominated all open-source solutions (including \effocr), but would have been orders of magnitude more costly to deploy. In our experience, it is frequently outside academic budgets for larger projects. 

Lightweight \effocr models are also faster than Tesseract and PaddleOCR - with the comparison to EasyOCR depending on the hardware used for deployment. This is despite having around 8x more parameters than Tesseract and around 4x more than EasyOCR (parameter count is similar to PaddleOCR). This is achieved through parallel rather than sequential decoding and ONNX integration. \effocr is also significantly more accurate on tasks like digitizing the 20 million U.S. historical newspaper scans.

Users for whom neither computational nor sample efficiency is of concern - because they are working in a well-resourced context and don't face cost constraints for the scale of their problem - are not our target audience and may well find an existing OCR engine like Google Cloud Vision better meets their needs. In practice, academic or large-scale archival digitization of document collections often involves low-resource languages or settings, tight budget constraints, or both.

\section{The \effocr Library}
\label{sec:core-library}

\subsection{Off-the-shelf Usage}

At the core of \effocr is an off-the-shelf toolkit. \effocr is a modular framework, that first localizes lines, characters, and (for some models) words using object detection, and then recognizes characters and words by embedding their crops and retrieving their nearest neighbor from an offline index of exemplar embeddings created from a digital font. 

\textbf{Localization:} \effocr supports two widely used backends for localization inference: MMDetection \cite{chen2019mmdetection}, which includes state-of-the-art object detection models, and Yolo \cite{yolov8}, which includes fast, efficient object detection models. Users can deploy line, word, and character models from the pre-trained model zoo, that use Yolo v8 \cite{yolov8} (optimized for efficiency), Yolo v5 \cite{Jocher_YOLOv5_by_Ultralytics_2020} (fewer dependencies) or Cascade R-CNN \cite{cai2018cascade} (optimized for accuracy). Pre-trained localization models are available for alphabetic English/Latin, Polytonic Greek, and CJK characters (which vary significantly in their aspect ratios and groupings). 

\textbf{Recognition:}
\effocr recognizes word and character crops using contrastively trained image retrieval models. 
The \effocr model zoo currently contains 30 pre-trained models, covering English, Polytonic Greek, and horizontally and vertically-written Japanese. 
We chose these languages to examine the utility of \effocr in a high resource setting, in a setting where existing solutions fail, and in an intermediate case. 

The \effocr pre-trained models use a variety of backbones: two lightweight convolutional backbones that are very efficient to deploy \cite{howard2019searching, maaz2022edgenext}, a state-of-the-art CNN encoder \cite{liu2022convnet}, and three vision transformers \cite{ali2021xcit, li2022exploring, liu2021swin}. 
For English, there is a word level model that defaults to character recognition when the word is below a default (tuneable) cosine similarity threshold, as well as a character-only model. 
%For Japanese, characters are words. 

The documentation provides more guidance on model selection. 
%As we show in Section \ref{sec:use-case}, \effocr has good few shot performance. We plan to crowd source a modest number of character/word crops for different languages from researchers who have requested that \effocr support their contexts off-the-shelf, combine these with digital fonts, and pre-train models with more scripts for a subsequent release, in addition to including user-contributed models.
A description of the training dataset is provided alongside with the trained models such that users can quickly identify the most suitable models for their tasks.

\effocr can be used off-the-shelf with just a few lines of code:

\begin{lstlisting}[language=python]
import effocr
engine = effocr.EffOCR(
    line_detector = "./line_model",
    localizer = "./localizer",
    char_recognizer = "./char_recognizer",
    word_recognizer = "./word_recognizer"
)
results = engine.infer('image.jpg')
\end{lstlisting}

ONNX \cite{onnxruntime} integration is an important component, as it allows for efficient CPU deployment and interoperability between deep learning frameworks. All \effocr stages can optionally employ ONNX-format models and ONNX-runtime inference, and models can be converted to ONNX format within the package. ONNX-runtime increases CPU throughput by up to four times \cite{Jocher_YOLOv5_by_Ultralytics_2020} for YOLO models used in \effocr, which allows for cost-effective cloud deployment for processing large document sets. ONNX compatibility allows additional model speedups through graph optimizations, quantization, and pruning. 

\subsection{Customized Model Training}
\label{sec:training}

Many low-resource settings are poorly served by existing OCR engines, and a central aim of EffOCR is to democratize OCR for these settings by providing a simple interface for custom model training that can be used by researchers and others who have limited experience with deep learning frameworks. 
Custom training can be initialized using a Yolo object detection model for localization and any timm image encoder model for recognition. In the near future, support for training localization models with MMDetection will be added. 
This futureproofs \effocr, as new models are developed. 

% \begin{lstlisting}[language=python]
% engine = effocr.EffOCR('config.yaml', 
%         'training_data.json', 
%         '/image_folder'
% )
% engine.train()
% \end{lstlisting}

\effocr supports logging of a training run on Weights and Biases \cite{wandb}.
It takes industry standard coco json labels as inputs, and hence is compatible with the outputs of a range of both open-source and proprietary labeling softwares. It also exports output in the same format, so that users can easily correct model predictions if desired to speed up labeling. 

Model training with \effocr is highly efficient, \textit{e.g.,} the convolutional backbones can be trained on Google Colab. 
We trained all models on either a single Nvidia RTX 3090 or A6000 card. 

\subsection{Visualization, Storage and Export}

\effocr comes with a tool to visualize the OCR, side-by-side with the original image, as well as to visualize the line, word, and character predictions. These greatly facilitate quality checking the output and troubleshooting potential problems. 

\effocr offers users different options for data export. The default outputs of \effocr include line coordinates, word coordinates, character coordinates, and the text associated with each of these annotations. The text for the full image is also assembled in the correct order. Users may choose to export only the assembled text, only text annotations associated with a given level of bounding box (line, word, or character), or all of the above. 

% \begin{lstlisting}[language=python]
% # Get only text
% results = engine.infer('image.jpg').text
% # Save all annotations
% results = engine.infer('image.jpg', save_coco_annotations='path/to/file.json')
% # Save word-level annotations
% results = engine.infer('image.jpg', save_coco_annotations='path/to/file.json', save_level='word')
% \end{lstlisting}

\subsection{User Contributions}
By making OCR sample efficient and easy to train, \effocr aims to promote the reusability and reproducibility of OCR pipelines. 
This is particularly important for low resource settings and languages, where there is little commercial incentive for product development and few alternatives to crowd-sourcing models. \effocr users can upload their self-trained models to the \effocr Hugging Face hub. Whenever a model is saved, a model card is automatically generated that follows best practices outlined in Hugging Face's Model Card Guidebook.\footnote{https://huggingface.co/docs/hub/model-card-guidebook} Moreover, the automatically generated card contains instructions on how to use the model in the context of \effocr and model-specific architecture and training details in the interest of reproducibility.

\subsection{Integration with Layout Parser}

OCR engines typically detect lines, versus detecting and classifying different layout objects in a document. Many documents have complex layouts - \textit{e.g.,} newspapers have headlines, articles, captions, ads, and headers arranged in complex multicolumn layouts, and tables likewise have different types of information arranged in oftentimes complex layouts. 
These structures necessitate applying object detection models for document layout analysis, which have been trained to detect the coordinates of each layout object and classify its type (\textit{e.g.,} headline, articles, etc). 

To facilitate combining \effocr with deep learning-based document layout models, 
wrappers will be integrated into a popular open-source layout detection package, Layout Parser \cite{shen2020large}, that will allow Layout Parser users to call any \effocr model. Layout Parser also has wrappers to call GCV and Tesseract, which will allow users to easily compare \effocr output to these other packages to decide what best meets their accuracy and cost objectives. Layout Parser and \effocr were designed by the same lab, facilitating long-run coordination between the packages.

\section{Applications}
\label{sec:use-case}

\begin{figure*}[ht]
    \centering
    \includegraphics[width=.65\linewidth]{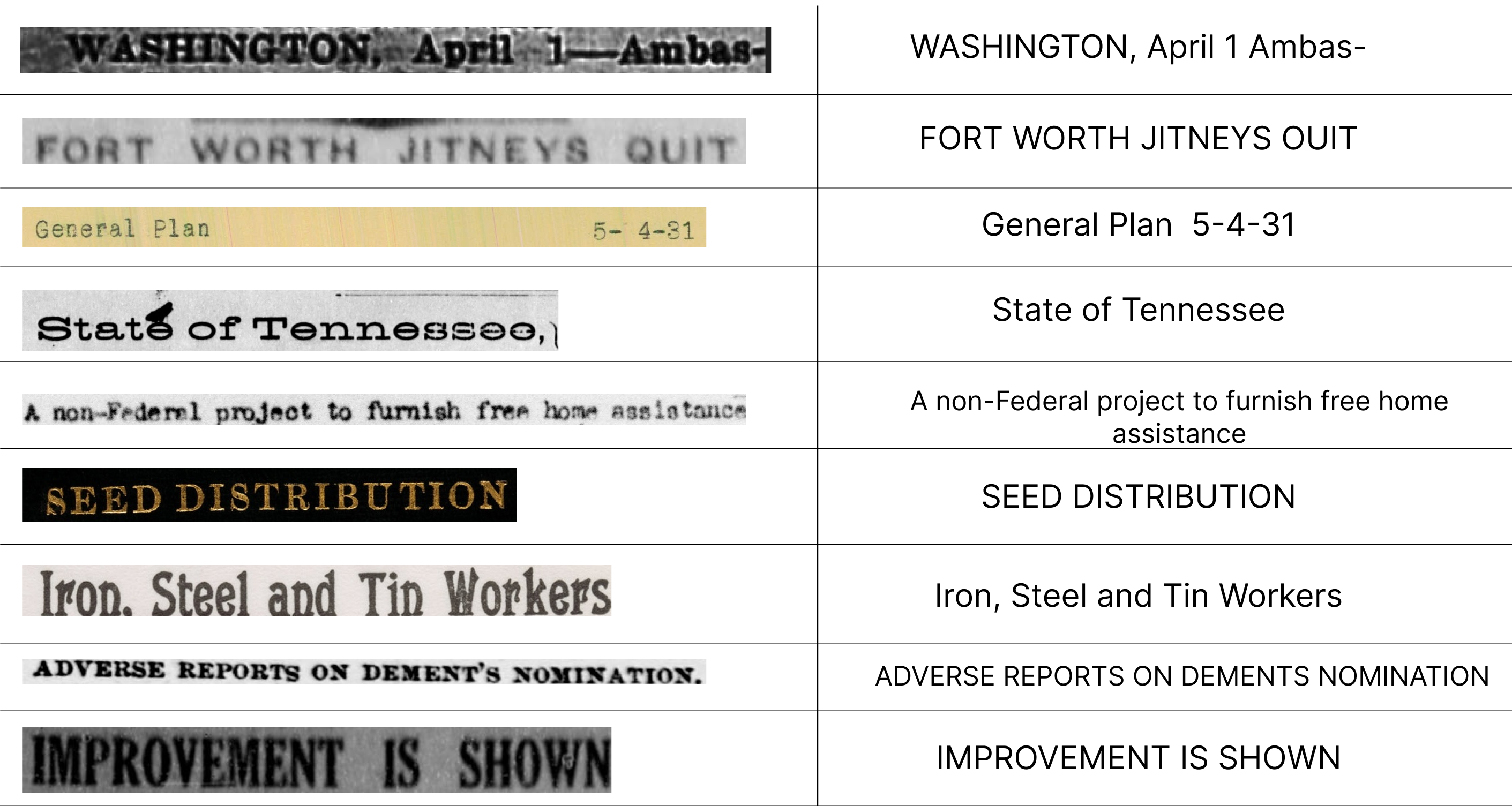}
    \caption{This figure shows a diversity of examples processed with \effocr, with predicted transcriptions on the right.}
    \label{fig:data}
    \vspace{-4mm}
  \end{figure*}

\textbf{Scalability:}
We have tested the utility of \effocr with various real-world applications. In the first application, we cheaply and accurately digitized 20 million newspaper page scans from Library of Congress's Chronicling America collection \cite{locca}. The resulting dataset, American Stories, is available for download on Hugging Face.\footnote{\url{https://huggingface.co/datasets/dell-research-harvard/AmericanStories}}.
Figure \ref{fig:data} illustrates why this is a challenging task: newspapers are extremely heterogeneous in their fonts and image quality. \citet{AmericanStories} provide a detailed analysis of the quality of the resulting text dataset.

We first trained character \effocr using synthetic data plus a labeled set of 291 newspaper lines \cite{carlson2023efficient}, created in a couple of hours. We then bootsrapped word level annotations by creating them with the character level \effocr model, filtering out lines with a high non-word rate. 

With \effocr, combined with layout analysis using Layout Parser, we could digitize the dataset with a \$60K USD cloud compute budget (plus pipeline development costs). GCV makes significant layout errors when fed full newspaper scans and achieves best performance when fed individual lines. At current prices, digitizing the collection at the line level, since GCV charges per image, would have cost over \$23 million USD. TrOCR Base, the most accurate open-source OCR, would have exceeded our budget by a factor of nearly 50. 

\textbf{Zero-Shot Performance:}
Second, we show that our English lightweight word-level model has strong zero-shot performance on randomly selected document collections from the U.S. National Archives. 
This model saw only newspapers in training, to test true zero shot performance.
We selected a single textline from each of 300 random documents from separate National Archive record groups. EffOCR achieved a 11.2\% CER on the diverse collection, compared with a 11.8\% CER from Tesseract (Best), a 12.1\% CER from EasyOCR, and a 51\% CER from TrOCR (Small), which appeared to struggle with blurry and partially obscured text. We suspect the results could be significantly improved by including a random sample of documents from the National Archives in training, to broaden the set of real world documents that the model is exposed to.

All open-source models performed significantly worse than GCV (1.2\% CER), but as discussed earlier cost concerns presently preclude its use at scale. Despite being engineered for low-resource, few-shot contexts, EffOCR remains competitive in high-resource, zero-shot situations. 

\textbf{Low Resource Settings:}
Finally, we use \effocr to digitize historical Japanese firm level records for vertically written Japanese documents \cite{teikoku}, where the best available solution (from Baidu OCR) mispredicts over half of characters. 
We use the evaluation set in \citet{carlson2023efficient}, which consists of randomly selected segments that were double labeled. 

Using a training set of 898 labeled table cells, we achieve a CER of 0.7\%, 80 times more accurate than the best existing solution. As a result, we are able to study a variety of questions about Japan's remarkable growth performance that would have been impossible to examine without \effocr. 

To further examine the limits of sample efficiency, we calculate the character classification error when the (character) model only sees one (or up to 5) labeled character(s) for each of the characters that appear in the training set, which comprise 77\% of the characters in the test set. This results in character classification errors of 13.4\% and 2.0\% respectively. While the model does clearly benefit from seeing multiple crops of characters that appear frequently, this illustrates viable few shot performance. 

\section{Limitations} \label{limits}

If large portions of a document are illegible, vision-only OCR will not be suitable and language understanding may be helpful for inferring content. For high resource languages such as English when cost is not a concern, users may get the best mileage from a leading commercial product such as GCV. 

Currently, the \effocr model zoo has pre-trained models supporting typewritten English, Japanese, and Polytonic Greek.  
Over the coming months, we will be crowd-sourcing annotations (including handwriting) from package users and colleagues. We will use them, along with digital fonts, to pre-train additional models. In addition, users are encouraged to contribute their models.

\effocr does not currently support handwriting. We started with typewritten documents because there are billions of public domain typeface documents that are of considerable interest to researchers and the general public. We are planning to expand the model zoo to include handwriting and users have already offered to contribute annotations. Synthetic handwriting generators, \textit{e.g.} \citet{bhunia2021handwriting}, can provide extensive data for pre-training for scripts that they support, analogous to the use of digital fonts for typeface documents. We will make synthetic handwriting datasets available so that package users can also use them for training their own custom models. 

%\bibliography{cites}
%\bibliographystyle{acl_natbib}

\clearpage

\setcounter{table}{0}
\renewcommand{\thetable}{S-\arabic{table}} % Setting the table number output to letters 
\setcounter{figure}{0}
\renewcommand{\thefigure}{S-\arabic{figure}} % Setting the figure number output to letters 
\setcounter{section}{0}
\renewcommand{\thesection}{S-\arabic{section}}

\onecolumn

\begin{center}
    \section*{Supplementary Materials}

\end{center}

\section{Model Architecture and Model Zoo}
Figure \ref{fig:arch} shows the \effocr model architecture, and Table \ref{model_zoo} summarizes the models in the \effocr model zoo.
Readers seeking technical details for the \effocr models contained in the pre-trained model zoo are referred to the detailed supplementary materials in \citet{carlson2023efficient}. 

%\section{Sample Efficiency}
\section{Sample Efficiency}
To examine how efficiently EffOCR learns in comparison to leading open source architectures, we train different OCR models from scratch using varying amounts of annotated data. EffOCR-C (Base) is compared to SVTR (implemented via PaddleOCR) \cite{du2022svtr}, CRNN (implemented via EasyOCR) \cite{shi2016end}, and TrOCR \cite{li2021trocr}. 
All architectures are pre-trained from scratch on 8,000 synthetic text lines, starting from pre-trained checkpoints not customized for OCR when supported by the framework. They are then fine-tuned on the study's benchmark datasets, with varying train-test-validation splits: 70\%-15\%-15\%, 50\%-25\%-25\%, 20\%-40\%-40\%, 5\%-47.5\%-47.5\%, and 0\%-50\%-50\% (i.e., zero-shot).
These exercises are performed for the English newspaper character level models and horizontal Japanese, as vertical Japanese is not supported by the comparison architectures. 

Figure \ref{fig:efficiency} plots the percentage of the benchmark dataset used in training on the x-axis and the CER on the y-axis.
On just 99 labeled table cells for Japanese and 21 labeled rows for LoCCA (the 5\% train split), EffOCR's CER is only 5\% (Japanese) and 7\% (English), showing viable few shot performance.  
The other architectures remain unusable. 
EffOCR performs nearly as well using 20\% or training data as using 70\%, where it continues to outperform all other alternatives. %, 
This illustrates that its parsimonious architecture learns efficiently. 

\section{Training Config Details}
The EffOCR package exposes a wide variety of training options and hyperparameters to users. A few key elements are described here, readers looking for more details are directed to the package documentation. 

\bigskip

\textit{Recognizer Training Options:}
\begin{itemize}
    \itemsep0em 
    \item \texttt{timm\_model\_name} Model name from timm \cite{rw2019timm} package used as a base encoder for the recognizer.
    \item \texttt{render\_dict} Folder to store crop renders and gold training data locally.
    \item \texttt{font\_dir\_path} Local path to draw tff (font) files from, which are used to create character/word renders.
    \item \texttt{hns\_txt\_path} Local file path to draw hard negative samples from. Hard negative text files are created by default at the end of recognizer training. Most recognizer training applications use two stages, an initial run and a hard negative sampling run. 
    \item \texttt{latin\_suggested\_args} Uses default arguments for alphabetic writing systems such as Latin, Greek, and Cyrillic. 
\end{itemize}
In addition to these options, a wide variety of standard model training parameters are exposed, including learning rate, optimizer options, weight decay, batch size, device selection, and number of training epochs. 

\bigskip

\textit{Localizer Training Options:}
\begin{itemize}
\itemsep0em
    \item \texttt{vertical} Whether model should expect characters aligned horizontally (as in English and many Latin scripts) or vertically (as in many character-based scripts). 
    \item \texttt{no\_words} Detect only characters, not words. Recommended for languages without word groupings. 
    \item \texttt{iou\_thresh} Training and validation IOU threshold for character/word detection.
    \item \texttt{conf\_thresh} Training and validation confidence threshold for character/word detection.
\end{itemize}
As with the recognizer, other standard training parameters are exposed. In particular, adjusting the image input shape may be valuable for particularly long or short lines. 

Hyperparameters and training procedure used to generate models listed in the Model Zoo (Table \ref{model_zoo}) are listed in \citet{carlson2023efficient}.

\section{Visualization} Figure \ref{fig:viz} shows the \effocr visualization interface. 

\section{American Stories} Figure \ref{fig:time} plots the number of articles in the \texttt{American Stories} dataset, created with \effocr, across time. 

\section{The Public Domain} Table \ref{copyright} provides detailed information about the requirements for information published in the United States to be in the public domain, in order to give readers a better sense of these collections.

\section{Inference Speed}
EffOCR implements two features designed to increase computational efficiency. First, both localization and recognition inference is run in a multithreaded fashion, ensuring that compute resources are fully utilized. Second, EffOCR provides support for ONNX runtime and ONNX-format models, which provide up to a 3x speedup on a CPU compared to native PyTorch runtime \cite{onnxruntime}. GPUs are typically cost prohibitive for digitization at scale.

Table \ref{colab_inf} provides a comparison between EffOCR and other commonly used OCR frameworks' python implementations. It is important to note that these numbers - across softwares - can vary significantly dependending on the hardware resources available. All comparisons are made on four 2200 MHz CPU cores, selected to represent a plausible and relatively affordable research compute setup. EffOCR performance is competitive with other widely used frameworks, with EffOCR (Small) having the fastest performance. Tesseract \cite{tesseract} testing used the \texttt{pytesseract} package with default settings. EasyOCR \cite{EasyOCR} testing used the \texttt{easyocr} package with default English settings. PaddleOCR \cite{Paddle} testing used the \texttt{paddleocr} package with \texttt{use\_angle\_cls} option and default English settings. TrOCR \cite{trocr2023Github} testing used the \texttt{transformers} package implementation, with \texttt{trocr-base-printed} and \texttt{trocr-small-printed} models for Base and Small tests, respectively. EffOCR testing used default settings with pretrained ONNX English newspaper models from the model zoo.

\clearpage
%\section{Tables and Figures}

\input{architecture.tex}
\clearpage

\input{SampleEfficiency.tex}
\clearpage

\input{viz.tex}
\clearpage

\input{ca_time.tex}

\clearpage

\input{tables/model_zoo}

\clearpage

\input{PublicDomain}

\clearpage

\input{tables/colab_speeds}

\clearpage

\bibliography{cites}
\bibliographystyle{acl_natbib}

\end{document}

%% file: architecture.tex
\begin{figure*}[ht]
    \centering
    \includegraphics[width=\linewidth]{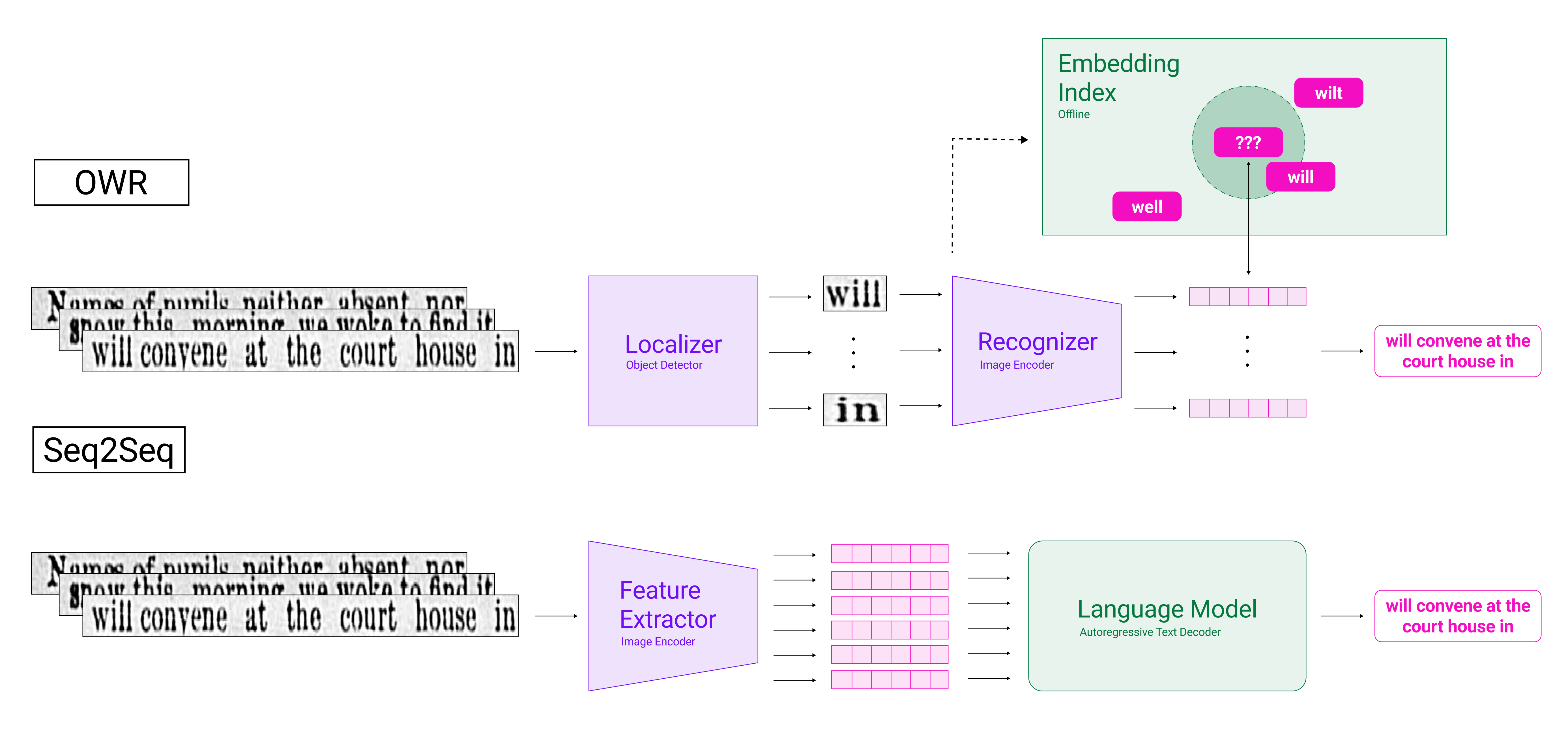}
    \caption{\textbf{EffOCR and Seq2Seq Model Architectures.} This figure represents the EffOCR architecture, as compared to a typical sequence-to-sequence OCR architecture.} 
    \label{fig:arch}
    \vspace{-4mm}
  \end{figure*}
  

%% file: SampleEfficiency.tex
\begin{figure*}[ht]
    \centering
    \includegraphics[width=\linewidth]{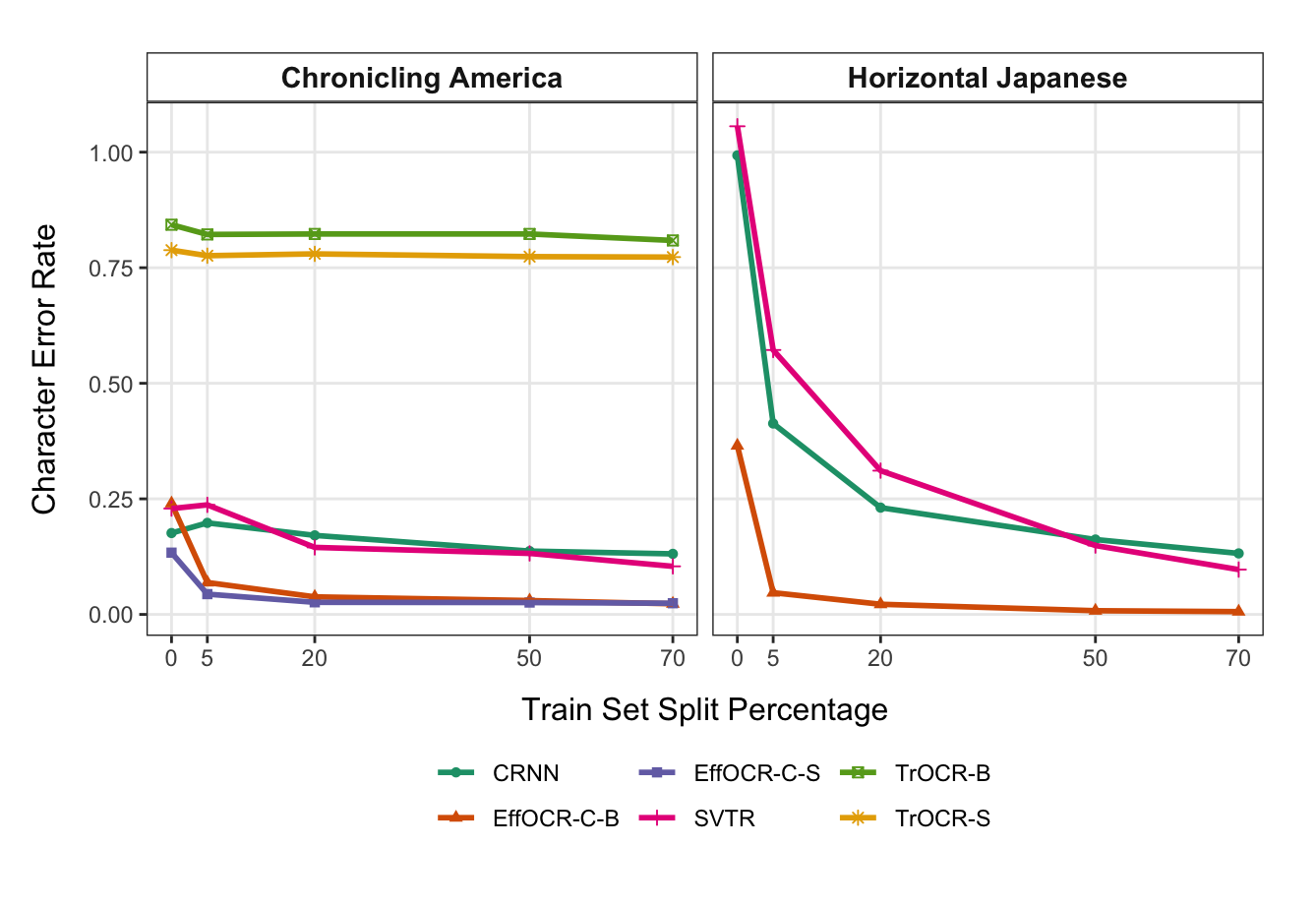}
    \caption{\textbf{Sample Efficiency.} This figure plots the percentage of the benchmark dataset used in training against the character error rate, for different OCR model architectures: CRNN (EasyOCR), SVTR (PaddleOCR), TrOCR (Transformer OCR), and \effocr small and base convolutional models.} 
    \label{fig:efficiency}
    \vspace{-4mm}
  \end{figure*}
  

%% file: viz.tex
\begin{figure*}[ht]
    \centering
    \includegraphics[width=\linewidth]{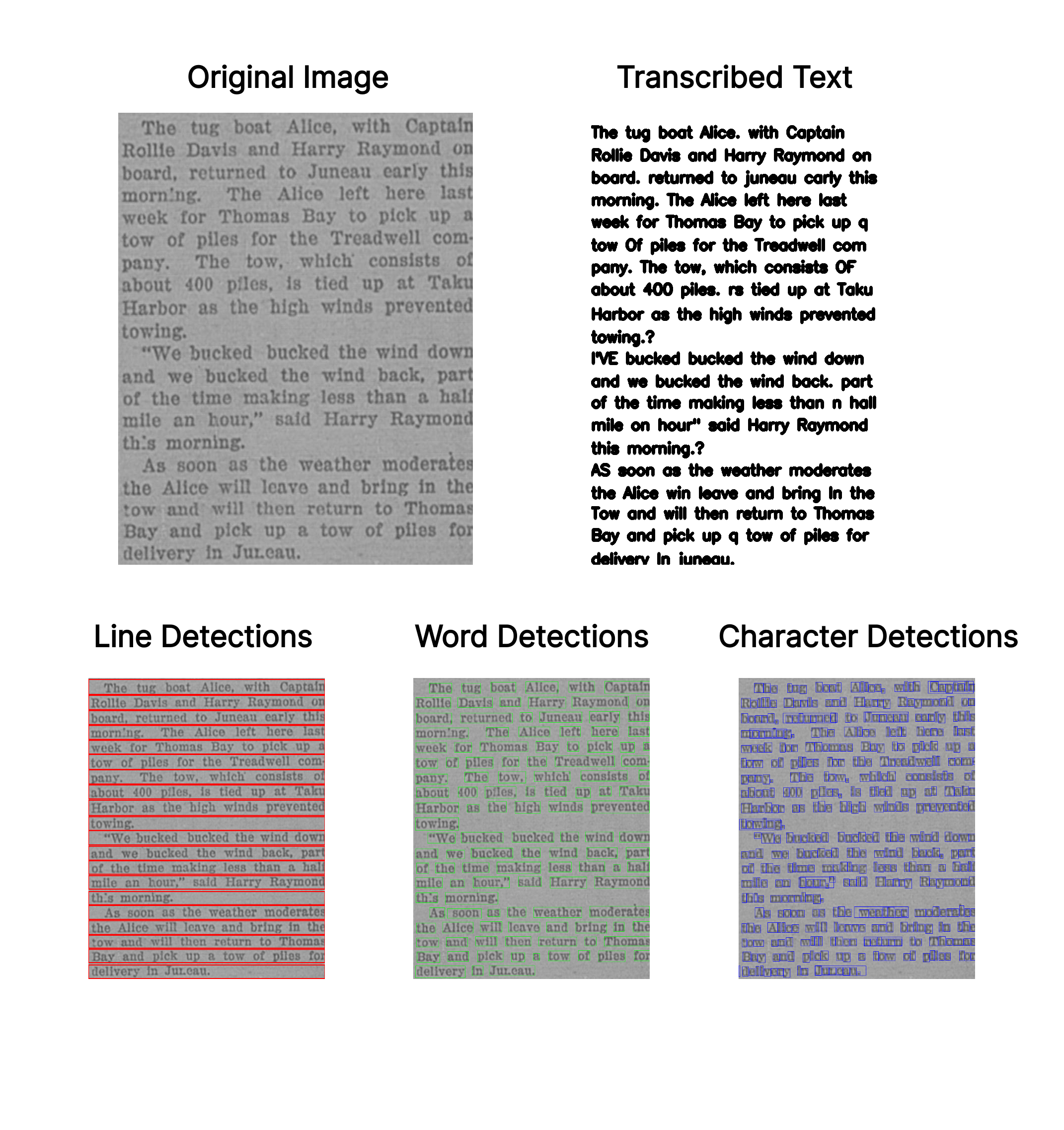}
    \caption{\textbf{Visualization.} This figure shows the \effocr visualization interface.} 
    \label{fig:viz}
    \vspace{-4mm}
  \end{figure*}
  

%% file: ca_time.tex
\begin{figure*}[ht]
    \centering
    \includegraphics[width=\linewidth]{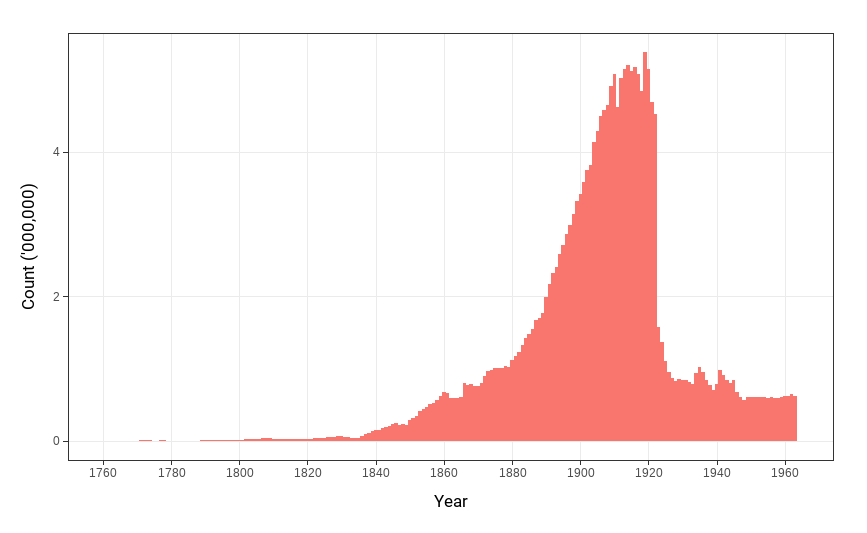}
    \caption{\textbf{American Stories.} This figure plots the number of articles in the \texttt{American Stories} dataset, created with \effocr, across time.} 
    \label{fig:time}
    \vspace{-4mm}
  \end{figure*}

%% file: tables/model_zoo.tex
\begin{table*}[ht]
    \centering
    \resizebox{\linewidth}{!}{
    \begin{threeparttable}
       \begin{tabular}{lcccccccccc}
      \toprule
\textbf{Training Set} &  \textbf{Line Detection}  &
\multicolumn{2}{c}{\textbf{Localizer}} & \multicolumn{2}{c}{\textbf{Word Recognition}} & \multicolumn{5}{c}{\textbf{Character Recognition}}\\

 &		YOLO	& 	YOLO & MaskRCNN	&	MobileNetV3	&	EdgeNeXt	&	MobileNetV3	&	EdgeNeXt & ViT & ConvNeXt & XCiT  \\
\cmidrule{1-1} \cmidrule(l{3pt}r{3pt}){2-2} \cmidrule(l{3pt}r{3pt}){3-4} \cmidrule(l{3pt}r{3pt}){5-6}  \cmidrule(l{3pt}r{3pt}){7-11} \\
English Newspapers	& 	\checkmark	&	\checkmark	&	\checkmark &	\checkmark	&	\checkmark	&	\checkmark	&	\checkmark	 & \checkmark & \checkmark & \checkmark \\
English Mixed Archival	 &		\checkmark	&	-	&	\checkmark  & \checkmark	&	-	&	\checkmark	& - & - & - & - \\
Japanese Vertical	 &		\checkmark	& \checkmark	&	\checkmark 	&  N/A &	N/A 	&	\checkmark	&	\checkmark	&	\checkmark & \checkmark & \checkmark	\\
Japanese Horizontal	 &	\checkmark	& \checkmark	& \checkmark	&	N/A	& N/A &	\checkmark	&	\checkmark	& \checkmark & \checkmark & \checkmark	  \\
Polytonic Greek	 &	\checkmark	& \checkmark	& -	&	N/A	& N/A &	\checkmark	&	-	& - & \checkmark & -	  \\
      \bottomrule 
    \end{tabular}
    \end{threeparttable}
  }
   \caption{ \raggedright Models Currently Available in the EffOCR Model Zoo. Note Japanese models do not use word-level recognition. }
      \label{model_zoo}
\end{table*}

%% file: PublicDomain.tex
\begin{table*}[htbp]
    \centering
    \resizebox{\linewidth}{!}{
    \begin{threeparttable}
       \begin{tabular}{lcc}
      \toprule
Date of Publication	&	Conditions	&	Copyright Term \\
\cmidrule{1-3}
\textbf{\textit{Public Domain}} \\
 \textbf{Anytime}	&	\textbf{Works prepared by an officer/employee of the}	&	\textbf{None} \\ 
 & \textbf{U.S. Government as part of their official duties} \\ \\
\textbf{Before 1928}	&	\textbf{None}	&	\textbf{None. Copyright expired.}	\\ \\
\textbf{1928 through 1977}	&	\textbf{Published without a copyright notice}	&	\textbf{None. Failure to comply with required formalities}	\\ \\
\textbf{1978 to 1 March 1989}	&	\textbf{Published without notice and }	&	\textbf{None. Failure to comply with required formalities}	\\
 & \textbf{without subsequent registration within 5 years} \\ \\
\textbf{1928 through 1963}	&	\textbf{Published with notice}	&	\textbf{None. Copyright expired}	\\
 & \textbf{but copyright was not renewed} \\ \\

\textbf{\textit{Copyrighted}} \\
1978 to 1 March 1989	&	Published without notice, but with 	&	70 (95) years after the death of author (corporate author)	\\
 & subsequent registration within 5 years \\ \\
1928 through 1963	&	Published with notice 	&	95 years after publication	\\
 & and the copyright was renewed \\ \\
1964 through 1977	&	Published with notice	&	95 years after publication 	\\ \\
1978 to 1 March 1989	&	Created after 1977 and published with notice	&	70 (95) years after the death of author (corporate author) \\
 & & or 120 years after creation, if earlier	\\ \\
1978 to 1 March 1989	&	Created before 1978 and first published 	&	The greater of the term specified in the previous entry \\ 
& with notice in the specified period & or 31 December 2047	\\ \\
From 1 March 1989 through 2002	&	Created after 1977	&	70 (95) years after the death of author (corporate author) \\
 & & or 120 years after creation, if earlier	\\ \\
From 1 March 1989 through 2002	&	Created before 1978 and &	The greater of the term specified in the previous entry \\ 
& first published in this period & or 31 December 2047	\\ \\
After 2002	&	None	&	70 (95) years after the death of author (corporate author) \\
& & or 120 years after creation, if earlier \\ \\

     \bottomrule 
    \end{tabular}
    \end{threeparttable}}
    \caption{This table summarizes U.S. copyright law, based on a similar table produced by the Cornell libraries. For concision, we focus on works initially published in the United States. A variety of other cases are also covered at \url{https://guides.library.cornell.edu/copyright}.}
      \label{copyright}
\end{table*}

%% file: tables/colab_speeds.tex
\begin{table}[ht]
\centering
\begin{tabular}{|l|c|c|}
\hline
Model & Textline/s & Article/s \\
\hline
EffOCR Base & 0.46 & 0.02 \\
EffOCR Small & 21.07 & 1.08 \\
Tesseract & 4.47 & 0.21 \\
EasyOCR & 19.80 & 1.03 \\
PaddleOCR & 13.56 & 0.61 \\
TrOCR (Base) & 0.43 & 0.02 \\
TrOCR (Small) & 0.97 & 0.05 \\
\hline
\end{tabular}
\caption{\raggedright Comparison of EffOCR speeds with other popular OCR frameworks in CPU environment. Tests included both Textline (single lines of text) and Article (5-40 lines of text) examples.}
\label{colab_inf}
\end{table}